\documentclass[10pt,twocolumn,letterpaper]{article}
\usepackage[table,xcdraw]{xcolor}
\usepackage{iccv}
\usepackage{times}
\usepackage{epsfig}
\usepackage{graphicx}
\usepackage{amsmath}
\usepackage{breqn}
\usepackage{amssymb}
\usepackage{epstopdf}
\usepackage{booktabs}
\usepackage{color}
\usepackage{comment}
\usepackage{float}
\usepackage{svg}
\usepackage{subfig}
\usepackage{mathtools}
\usepackage{wasysym}
\usepackage{multirow}
\usepackage[toc,page]{appendix}

\usepackage[pagebackref=true,breaklinks=true,letterpaper=true,colorlinks,bookmarks=false]{hyperref}

\iccvfinalcopy 

\begin{document}

\title{Chained Multi-stream Networks Exploiting Pose, Motion, and Appearance for Action Classification and Detection}

\author{Mohammadreza Zolfaghari , Gabriel L. Oliveira, Nima Sedaghat, and Thomas Brox \\
University of Freiburg\\
Freiburg im Breisgau, Germany\\
{\tt\small \{zolfagha,oliveira,nima,brox\}@cs.uni-freiburg.de}
}

\maketitle

\begin{abstract}
General human action recognition requires understanding of various visual cues. In this paper, we propose a network architecture that computes and integrates the most important visual cues for action recognition: pose, motion, and the raw images. For the integration, we introduce a Markov chain model which adds cues successively. The resulting approach is efficient and applicable to action classification as well as to spatial and temporal action localization. The two contributions clearly improve the performance over respective baselines. 
The overall approach achieves state-of-the-art action classification performance on HMDB51, J-HMDB and NTU RGB+D datasets. Moreover, it yields state-of-the-art spatio-temporal action localization results on UCF101 and J-HMDB.
\end{abstract}

\section{Introduction}
Human action recognition is a complex task in computer vision, due to the variety of possible actions is large and there are multiple visual cues that play an important role. In contrast to object recognition, action recognition involves not only the detection of one or multiple persons, but also the awareness of other objects, potentially involved in the action, such as the pose of the person, and their motion. Actions can span various time intervals, making good use of videos and their temporal context is a prerequisite for solving the task to its full extent \cite{comp_ltc,Tran2015c3d}.

\begin{figure}[t]
\centering
\includegraphics[width=0.48\textwidth]{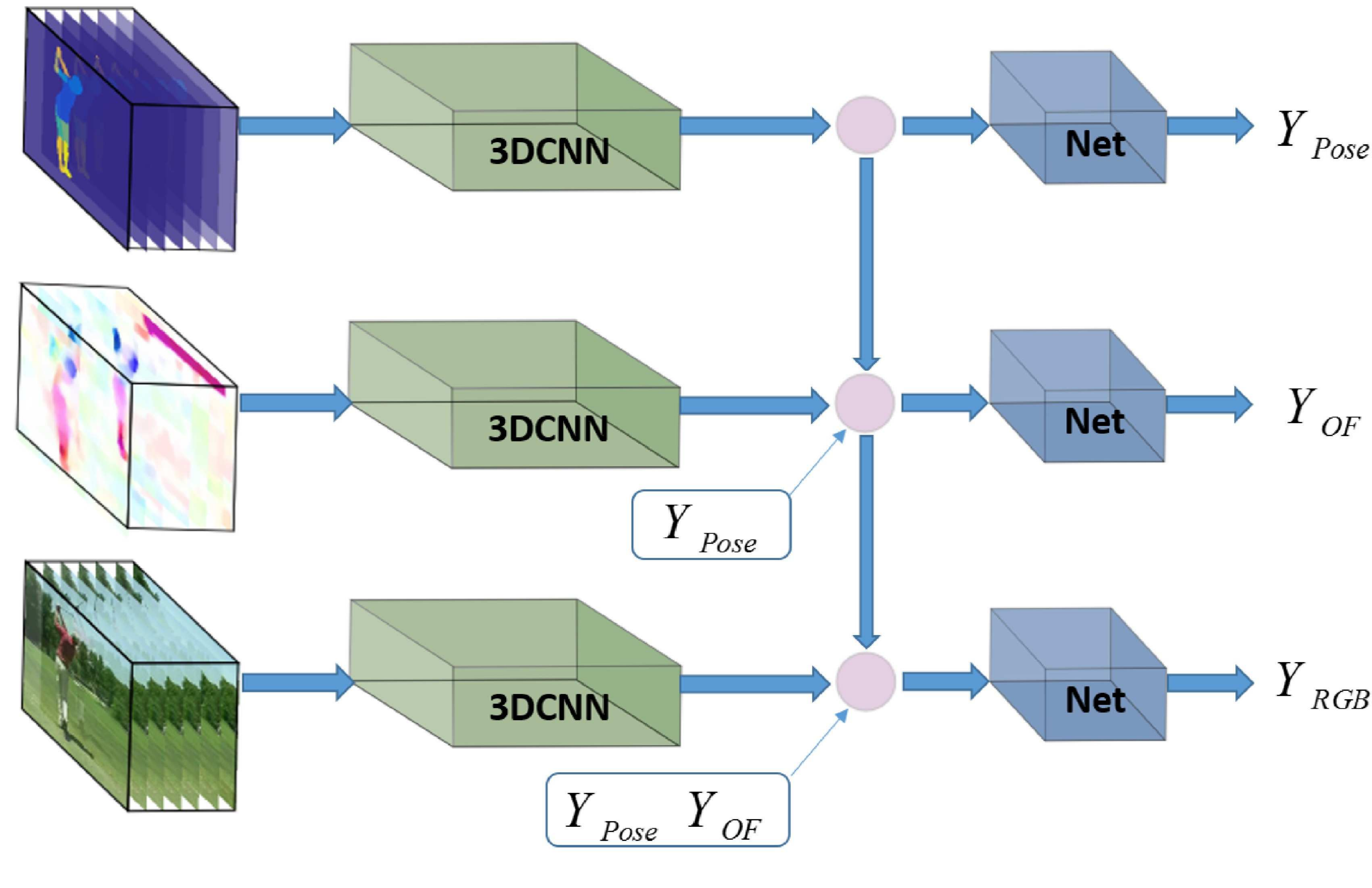}
\caption{The chained multi-stream 3D-CNN sequentially refines action class labels by analyzing motion and pose cues. Pose is represented by human body parts detected by a deep network. The spatio-temporal CNN can capture the temporal dynamics of pose. Additional losses on $Y_{Pose}$ and $Y_{OF}$ are used for training. The final output of the network $Y_{RGB}$ is provided at the end of the chain.}
\label{covergirl}
\end{figure}

The success of convolutional networks in recognition has also influenced action recognition.
Due to the importance of multiple visual cues, as shown by Jhuang et al.~\cite{JhmdbDataset}, multi-stream architectures have been most popular. This trend was initiated by Simonyan and Zisserman \cite{comp_zisserman}, who proposed a simple fusion of the action class scores obtained with two separate convolutional networks, where one was trained on raw images and the other on optical flow. The relative success of this strategy shows that deep networks for action recognition cannot directly infer the relevant motion cues from the raw images, although, in principle, the network could learn to compute such cues.

In this paper, we propose a three-stream architecture that also includes pose, see Figure \ref{covergirl}. Existing approaches model the temporal dynamics of human postures with hand-crafted features. We rather propose to compute the position of human body parts with a fast convolutional network. Moreover, we use a network architecture with spatio-temporal convolutions \cite{Tran2015c3d}. This combination can capture temporal dynamics of body parts over time, which is valuable to improve action recognition performance, as we show in dedicated experiments. 
The pose network also yields the spatial localization of the persons, which allows us to apply the approach to spatial action localization in a straightforward manner. 

The second contribution is on the combination of the multiple streams, as also illustrated in Figure \ref{covergirl}. The combination is typically done by summation of scores, by a linear classifier, or by early or late concatenation of features within the network. In this paper, we propose the integration of different modalities via a Markov chain, which leads to a sequential refinement of action labels. We show that such sequential refinement is beneficial over independent training of streams. At the same time, the sequential chain imposes an implicit regularization. This makes the architecture more robust to over-fitting -- a major concern when jointly training very large networks. Experiments on multiple benchmarks consistently show the benefit of the sequential refinement approach over alternative fusion strategies.  

Since actions may span different temporal resolutions, we analyze videos at multiple temporal scales. We demonstrate that combining multiple temporal granularity levels improves the capability of recognizing different actions. In contrast to some other state-of-the-art strategies to analyze videos over longer time spans, e.g., temporal segmentation networks \cite{comp_tsn}, the architecture still allows the temporal localization of actions by providing actionness scores of frames using a sliding window over video. We demonstrate this flexibility by applying the approach also to temporal and spatio-temporal action detection. Compared to previous spatio-temporal action localization methods, which are typically based on region proposals and action tubes, the pose network in our approach directly provides an accurate person localization at no additional computational costs. Therefore, it consistently outperforms the previous methods in terms of speed and mean average precision.

\section{Related work}
\label{sec:relatedwork}

\textbf{Feature based approaches.}
Many traditional works in the field of action recognition focused on designing features to discriminate action classes \cite{rw_localfeature,comp_idt,rw_hog,rw_stip}.
%
These features were encoded with high order encodings, e.g., bag of words (BoW) \cite{rw_bow} or Fisher vector based encodings \cite{rw_fv}, to produce a global representation for video and to train a classifier on the action labels.
Recent research showed that most of these approaches are not only computationally expensive, but they also fail on capturing context and high-level information.

\begin{figure*}
\centering
\includegraphics[width=0.98\textwidth]{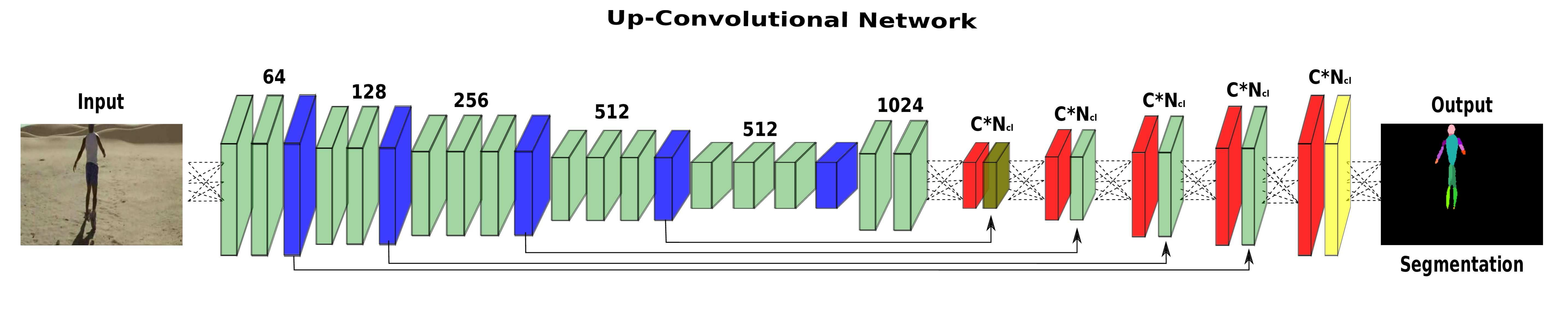}
\caption{Human body part segmentation architecture. Convolutions are shown in green, pooling in blue, feature map dropout in brown, up-convolutional layers in red and softmax in yellow.}
\label{parts_arch}
\end{figure*}

\textbf{CNN based approaches.}
Deep learning has enabled the replacement of hand-crafted features by learned features, and the learning of whole tasks end-to-end. Several works employed deep architectures for video classification \cite{comp_glstm, Tran2015c3d, comp_tdd}. Thanks to their hiearchical feature representation, deep networks learn to capture localized features as well as context cues and can exploit high-level information from large scale video datasets.  Baccouche et al. \cite{Baccouche11} firstly used a 3D CNN to learn spatio-temporal features from video and in the next step they employed an LSTM to classify video sequences. 
More recently, several CNN based works presented efficient deep models for action recognition \cite{comp_et3dcnn,comp_tsrcnn,Tran2015c3d}. Tran et al.  \cite{Tran2015c3d} employed a 3D architecture to learn spatio-temporal features from videos. 

\textbf{Fusion of multiple modalities.} Zisserman et al. \cite{comp_zisserman} proposed a two-stream CNN to capture the complementary information from appearance and motion, each modality in an independent stream. Feichtenhofer et al. \cite{comp_twofusion} investigated the optimal position within a convolution network in detail to combine the separate streams. Park et al.~\cite{rw_comb} proposed a gated fusion approach. In a similar spirit, Wang et al.~\cite{rw_audio} presented an adaptive fusion approach, which uses two regularization terms to learn fusion weights.
In addition to optical flow, some works made use of other modalities like audio \cite{rw_audio}, warped flow \cite{comp_tsn}, and object information \cite{rw_object} to capture complementary information for video classification. 
In the present work, we introduce a new, flexible fusion technique for early or late fusion via a Markov chain and show that it outperforms previous fusion methods. 

\textbf{Pose feature based methods.}
Temporal dynamics of body parts over time provides strong information on the performing action. Thus, this information has been employed for action recognition in several works \cite{comp_pcnn,rw_hpose,rw_pose_chu}. Cheron et al. \cite{comp_pcnn} used pose information to extract high-level features from appearance and optical flow. They showed that using pose information for video classification is highly effective. Wang et al.~\cite{rw_pose_chu} used data mining techniques to obtain a representation for each video and finally, by using a bag-of-words model to classify videos. 
In the present work, we compute the human body layout efficiently with a deep network and learn the relevant spatio-temporal pose features within one of the streams of our action classification network.

\section{Inputs to the Network}

We rely on three input cues: the raw RGB images, optical flow, and human pose in the form of human body part segmentation. All inputs are provided as spatio-temporal inputs covering multiple frames. 

\subsection{Optical Flow}

We compute the optical flow with the method from Zach et al.~\cite{flow_tvl1}, which is a reliable variational method that runs sufficiently fast. We convert the x-component and y-component of the optical flow to a 3 channel RGB image by stacking components and magnitude of them \cite{comp_tsrcnn}. The flow and magnitude values in the image are multiplied by $16$ and quantized into the [0,255] interval \cite{comp_granular, comp_tsrcnn, cvpr16_detect, comp_tsn}.

\subsection{Body Part Segmentation}

Encoder-decoder architectures with an up-convolutional part have been used successfully for semantic segmentation tasks \cite{long_shelhamer_fcn,liu2015,Ronneberger2015,kendall2015,Oliveira2016}, depth estimation \cite{LiuDepth} and optical flow estimation~\cite{FlowNet}.
For this work, we make use of Fast-Net \cite{Oliveira2016}, a network for human body part segmentation, which will provide our action recognition network with body pose information. Figure \ref{parts_arch} illustrates the architecture of Fast-Net. The encoder part of the network is initialized with the VGG network~\cite{VGG}. Skip connections from the encoder to the decoder part ensure the reconstruction of details in the output up to the original input resolution.

\begin{figure}[t]
\centering
\includegraphics[width=0.46\textwidth]{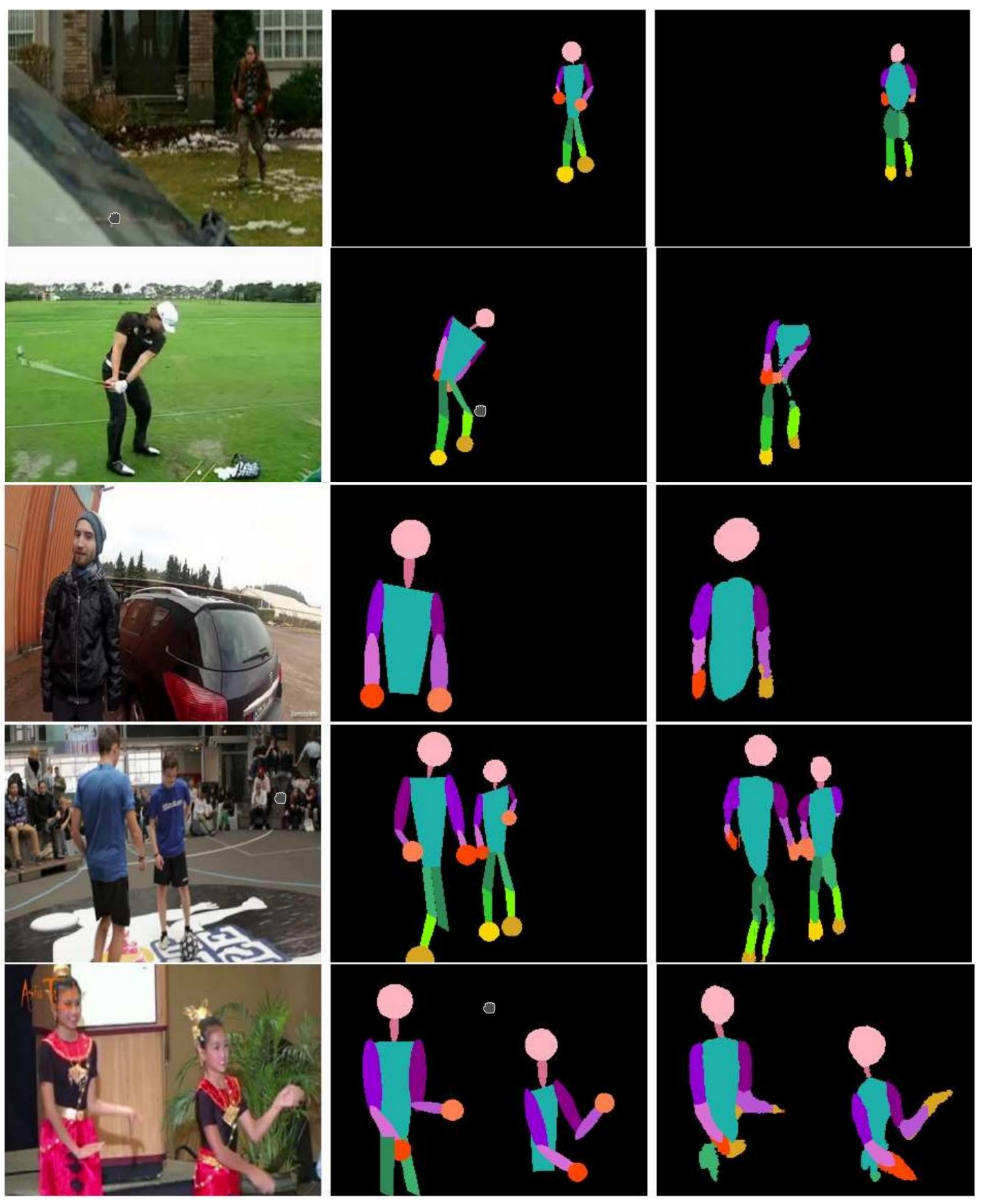}
    \caption{Qualitative results on J-HMDB and MPII datasets (task with 15 body 
parts). \textbf{First column:} Input image. \textbf{Second column:} Ground truth. 
\textbf{Third column:} Result predicted with Fast-Net. First two rows correspond to results on J-HMDB and the last ones on MPII.}
    \label{fig:QualitativePose}
\end{figure}

We trained the Fast-Net architecture on the J-HMDB \cite{JhmdbDataset} and the MPII \cite{mpiiDataset} action recognition datasets. J-HMDB provides body part segmentation masks and joint locations, while MPII provides only joint locations. To make body part masks compatible across datasets, we apply the following methodology, which only requires annotation for the joint locations. 
First, we derive a polygon for the torso from the joint locations around that area. Secondly, we approximate the other parts by ellipses scaled consistently based on the torso area and the distance between the respective joints; see second column of Fig. \ref{fig:QualitativePose}.     
We convert the body part segmentation into a 3 channel RGB image, mapping each label to a correspondent pre-defined RGB value.

To the best of our knowledge, we are the first who trained a convolutional network on body part segmentation for the purpose of action recognition. 
Figure \ref{fig:QualitativePose} shows exemplary results of the body part segmentation technique on J-HMDB and MPII datasets. 
Clearly, the network provides good accuracy on part segmentation and is capable of handling images with multiple instances. The pose estimation network has a resolution of 150$\times$150 and runs at $33$ fps. 

\section{Action Recognition Network}

\subsection{Multi-stream Fusion with a Markov Chain}

To integrate information from the different inputs we rely on the model of a multi-stream architecture~\cite{comp_zisserman}, i.e., each input cue is fed to a separate convolutional network stream that is trained on action classification. The innovation in our approach is the way we combine these streams. In contrast to the previous works, we combine features from the different streams sequentially. Starting with the human body part stream, we refine the evidence for an action class with the optical flow stream, and finally apply a refinement by the RGB stream. 

We use the assumption that the class predictions  are conditionally independent due to the different input modalities. Consequently, the joint probability over all input streams factorizes into the conditional probabilities over the separate input streams. 



\begin{figure}[t]
\centering
    \subfloat{%
       \includegraphics[width=.47\linewidth]{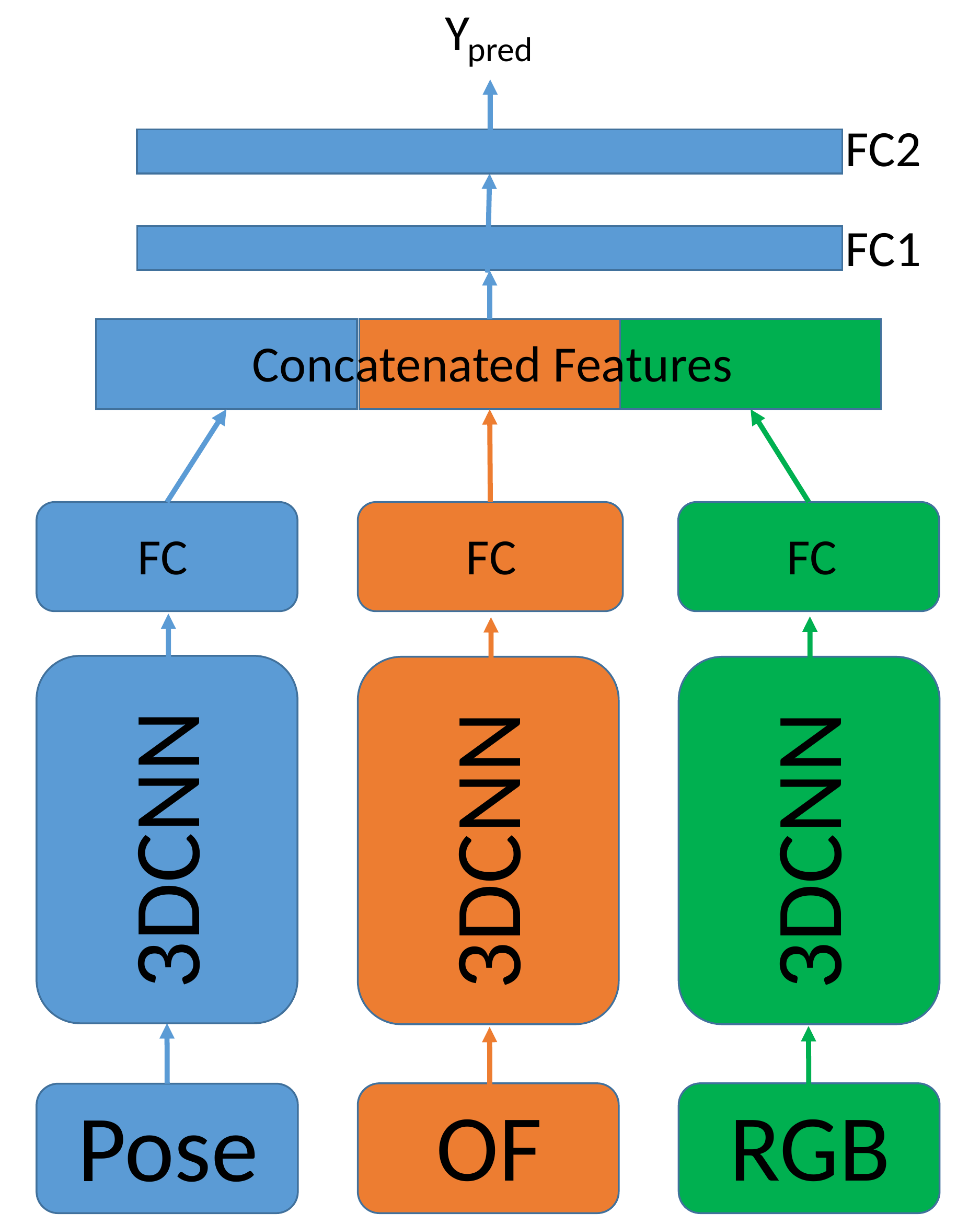}
        \includegraphics[width=.47\linewidth]{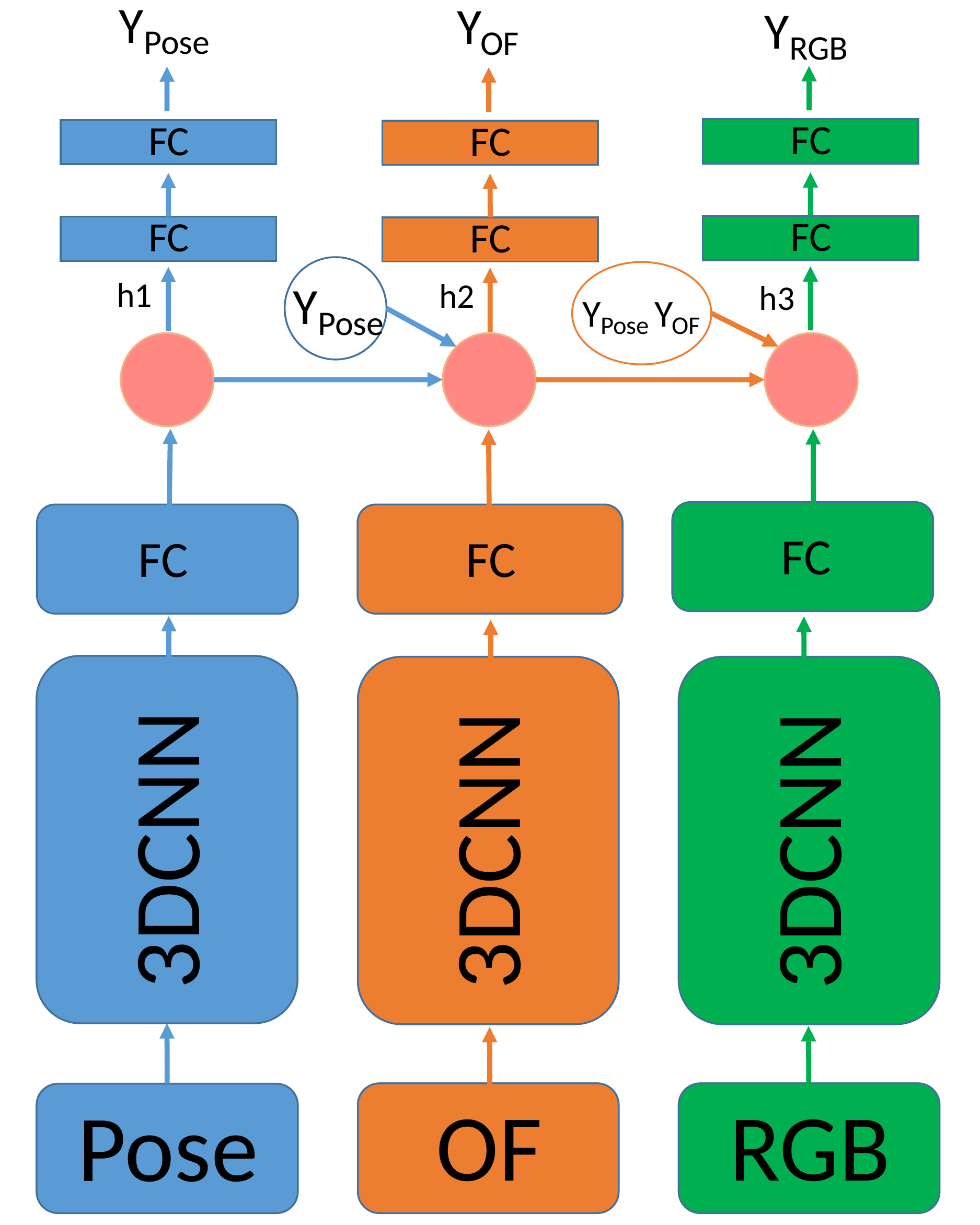}
        
    }
\caption{Baseline fusion architecture (left) and the proposed approach (right). In the chained architecture, there is a separate loss function for each stream. The final class label is obtained at the end of the chain (rightmost prediction).}
\label{baselinefig}
\end{figure}

In a Markov chain, given a sequence of inputs $X=\{X_1,X_2,...,X_S\}$, we wish to predict the output sequence $Y=\{Y_1,Y_2,...,Y_S\}$ such that $P(Y|X)$ is maximized. Due to the Markov property, $P(Y|X)$ can be decomposed: 
\begin{multline}
\label{eq1}
P(Y|X)=P(Y_1|X)\prod_{s=2}^{S}P(Y_s|X,Y_1,\dots,Y_{s-1})
\end{multline}
For the state $s\in \{1,\dots,S\}$, we denote by $h_s$ the hidden state of that stream.
We use deep networks to model the likelihood in \eqref{eq1}:

\begin{figure*}
\centering
\includegraphics[width=0.96\textwidth]{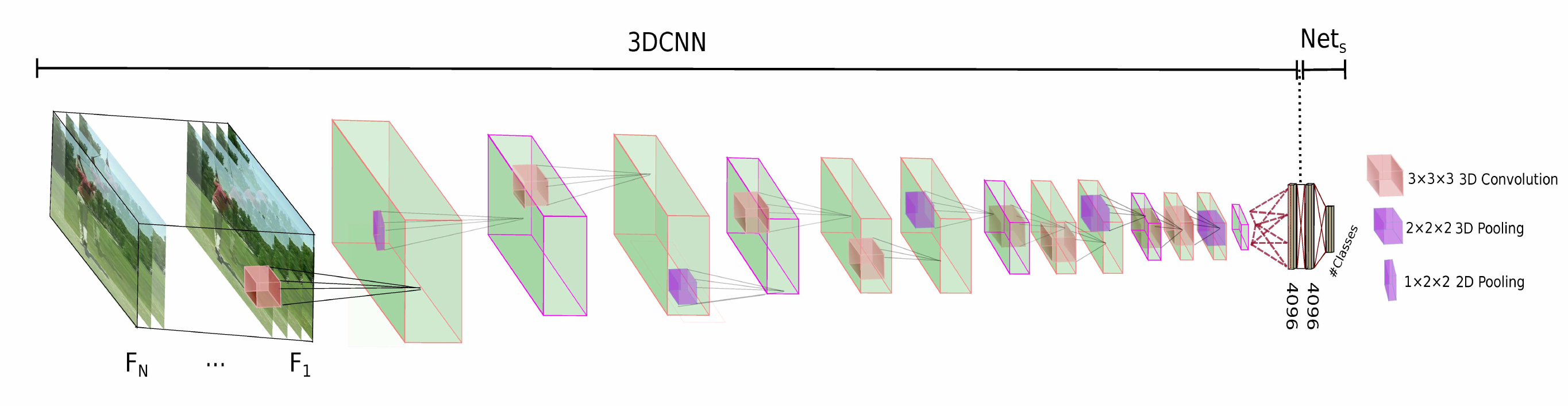}
\caption{Base architecture used in each stream of the action recognition network. The convolutional part is a 3DCNN architecture. We define the remaining fully connected layers as $Net_{s}$. }
\label{baseArchitecture}
\end{figure*}

\begin{equation}
\begin{aligned}
 &h_s =f([h_{s-1},{\rm 3DCNN}(X_s),(Y_1,\dots, Y_{s-1})]) \\
 &P(Y_s|X,Y_{<s}) = {\rm softmax}({\rm Net}_s(h_s)),
\end{aligned}
\end{equation}
where $f$ is a non-linearity unit (ReLU), $h_{s-1}$ denotes the hidden state from the previous stream, and $y_s$ is the prediction of stream $s$. For the ${\rm 3DCNN(\cdot)}$, we use the convolutional part of the network presented in Figure~\ref{baseArchitecture} to encapsulate the information in the input modality, and ${\rm Net}_s$ is the fully connected part in Figure~\ref{baseArchitecture}. 

At each fusion stage, we concatenate the output of the function ${\rm 3DCNN(\cdot)}$ with the hidden state and the outputs from the previous stream and apply the non-linearity $f$ before feeding them to ${\rm Net}_s$. Finally, at the output part, we use ${\rm Net}_s$ to predict action labels from $h_s$. With the ${\rm softmax(\cdot)}$ function we convert these scores into \mbox{(pseudo-)probabilities}.

Using the above notation, we consider input modalities as $X=\{X_{pose},X_{OF},X_{RGB}\}$, and $X_s=\{x_t\}^T_{t=1}$, where $x_t$ is the $t$-th frame in $X_s$, and $T$ is the total number of frames in $X_s$. At the stage $s=1$, by considering $X_{1}=X_{pose}$ we start with an initial hidden state and obtain an initial prediction (see Figure \ref{baselinefig}-right): 
\begin{equation}
\begin{aligned}
h_1 &={\rm 3DCNN}(X_{pose})\\
P(Y_1|X) &={\rm softmax}({\rm Net}_1(h_1))
\end{aligned}
\end{equation}
\\
At each subsequent stage $s\geqslant 2 $, we obtain a refined prediction $y_s$ by combining the hidden state and the predictions from the previous stage.


\begin{equation}
\begin{aligned}
 &h_2 =f([h_{1},{\rm 3DCNN}(X_{OF}),(Y_1)])\\
 &P(Y_2|X,Y_{<2}) ={\rm softmax}({\rm Net}_2(h_2))
 \\ \\
  &h_3 =f([h_{2},{\rm 3DCNN}(X_{RGB}),(Y_1,Y_{2})])\\
 &P(Y_3|X,Y_{<3}) ={\rm softmax}({\rm Net}_3(h_3))
\end{aligned}
\end{equation}

In the proposed model, at each stage, the next prediction is made conditioned on all previous predictions and the new input. Therefore, when training the network, the prediction of the output class label does not only depend on the input, but also on the previous state. Thus, the network in that stream will learn complementary features to refine the class labels from the previous streams. With this chaining and joint training, the information at the previous stages serve as the present belief for the predictions at the current stage, as shown in Figure \ref{baselinefig}-right. This sequential improvement of the class label enables the combination of multiple cues within a large network, while keeping the risk of over-fitting low. 

This is in contrast to the fusion approaches that combine features from different, independently trained streams. In such a case, the different streams are not enforced to learn complementary features. In the other extreme, approaches that train all streams jointly but not sequentially, are more prone to over-fitting, because the network is very large, and, in such case, lacks the regularization via the separate streams and their additional losses.

It should be expected that the ordering of the sequence plays a role for the final performance. We compared different ordering options in our experiments and report them in the following section. The ordering that starts with the pose as input and ends with the RGB image yielded the best results. 

It is worth noting that the concept of sequential fusion could be applied to any layer of the network. Here we placed the fusion after the first fully-connected layer, but the fusion could also be applied to the earlier convolutional layers.

\subsection{Network Configuration}

In all streams, we use the C3D architecture \cite{Tran2015c3d} as the base architecture, which has $17.5M$ parameters. The network has $8$ three-dimensional convolution layers with kernel size of $3\times3\times3$ and stride 1, $5$ three-dimensional pooling layers with kernel size of $2\times2\times2$ and stride 2 and two fully connected layers followed by a softmax; see Figure~\ref{baseArchitecture}. Each stream is connected with the next stream via layer FC$6$; see Figure~\ref{baselinefig}-right. Each stream takes 16 frames as input. 

\subsection{Training}

The network weights are learned using mini-batch stochastic gradient descent (SGD) with a momentum of $0.9$ and weight decay of $5e^{-4}$. We jointly optimize the whole network without truncating gradients and update the weights of each stream based on the full gradient including the contribution from the following stream. We initialize the learning rate with $1e^{-4}$ and decrease it by a factor of $10$ every $2k$ for J-HMDB, $20k$ for UCF101 and NTU, and at multiple steps for HMDB51.
The maximum number of iterations was $20k$ for J-HMDB, $40k$ for HMDB51 and $60k$ for the UCF101 and NTU datasets.
We initialize the weights of all streams with an RGB network pre-trained on the large-scale Sports-1M dataset \cite{Sports1M}.

We split each video into clips of $16$ frames with an overlap of $8$ frames and feed each clip individually into the network stream with size of $16\times112\times112$. 
We apply corner cropping as a form of data augmentation to the training data. Corner cropping extracts regions from the corners and the center of the image. It helps to prevent the network from bias towards the center area of the input. Finally, we resize these cropped regions to the size of $112\times112$.
In each iteration, all streams take the same clip from the video with the same augmentation but with different modalities as input.

We used Caffe \cite{jia2014caffe} and an NVIDIA Titan X GPU to run our experiments. 
The training time for the J-HMDB dataset was ${\sim 10}$ hours for the full network. 

\subsection{Temporal Processing of the Whole Video}

At test time, we feed the architecture with a temporal window of $16$ frames. The stride over the video is $8$. Each set of inputs is randomly selected for cropping operations, which are $4$ corners and $1$ center crop for the original image and their horizontal flipping counterpart. 
We extract scores before the softmax normalization in the last stream ($Y\_{RGB}$). 

In case of action classification, the final score of a video is calculated by taking the average of scores over all temporal windows across a video and 10 crop scores per clip. 
Apart from averaging, we also tested a multi-resolution approach, which we call \textbf{multi-granular (MG)}, where we trained separate networks for three different temporal resolutions. These are assembled as (1) $16$ consecutive frames, (2) $16$ frames from a temporal window of $32$ frames by a sample rate of $2$, and (3) $16$ frames sampled randomly from the entire video. For the final score, we take the average over the scores produced by these temporal resolution networks. This approach extends the temporal context that the network can see, which can be useful for more complex actions with longer duration. 

In case of temporal action detection, we localize the action in time by thresholding the score provided for each frame. Clearly, the MG approach is not applicable here. In addition to the action score, also the human body part network helps in temporal localization: we do not detect an action as long as no human is detected. More details on the spatio-temporal action detection are provided in the experimental section and in the supplemental material.

\section{Experiments}
\label{experiments}

\subsection{Datasets}

\textbf{UCF-101} \cite{UCF101} contains more than $2$ million frames in more than
$13,000$ videos, which are divided into $101$ human action classes. The dataset is split into three folds and each split contains about $8000$ videos for training. 
The UCF101 dataset also comes with a subset for spatio-temporal action detection. 

\textbf{HMDB51} \cite{HMDB51} contains $6766$ videos divided into $51$ action classes, each with at least $101$ samples. 
The evaluation follows the same protocol used for UCF-101.

\textbf{J-HMDB} contains a subset of videos from the HMDB dataset, for which it provides additional annotation, in particular optical flow and joint localization \cite{JhmdbDataset}. Thus, it is well-suited for evaluating the contribution of optical flow, body part segmentation, and the fusion of all cues via a Markov chain. 
The dataset comprises $21$ human actions. 
The complete dataset has $928$ clips and $31838$ frames. 
There are $3$ folds for training and testing for this dataset. 
The videos in J-HMDB are trimmed and come with bounding boxes. Thus, it can be used also as a benchmark for spatial action localization. 

\textbf{NTU RGB+D} is a recent action recognition dataset that is quite large and provides depth and pose ground truth \cite{comp_ntu_cvpr}. It contains more than $56000$ sequences and $4$ million frames. NTU provides $60$ action classes and $3D$ coordinates for $25$ joints. Additionally, the high intra-class variations make NTU one of the most challenging datasets.


\begin{table}
\small
\centering
\begin{tabular}{|c|l|c|c|c|}
\hline
\textbf{Streams} & \textbf{Variant} & \textbf{UCF101} & \textbf{HMDB} & \textbf{J-HMDB} \\ \hline
\multirow{4}{*}{1}   & RGB                 & 84.2\%   & 53.3\% & 60.8\% \\ 
                     & OF                & 79.6\%   & 45.2\% & 61.9\% \\ 
                     & Pose              & 56.9\%   & 36.0\% & 45.5\% \\ 
                     & Pose (GT)         & -        & -      & 56.8\% \\ \hline 
\multirow{3}{*}{RGB+OF} 

 & baseline                              & 87.1\%   & 55.6\% & 62.7\% \\ 
 & chained                               & 88.9\%   & 61.7\% & 72.8\% \\ 
 & chained+MG                            & -        & 66.0\% & -      \\ \hline
\multirow{3}{*}{3 w/o GT}
 & baseline                              & 89.1\%   & 57.5\% & 70.2\% \\
 & chained                               & 90.4\%   & 62.1\% & 79.1\% \\
 & chained+MG                            & \bf{91.3\%}   & \bf{71.1\%} & -      \\ \hline
\multirow{2}{*}{3 with GT} 
 & baseline                              & -        & -      & 72.0\% \\
 & chained                               & -        & -      & \bf{83.2\%} \\  \hline
\end{tabular} \vspace{0mm}
\caption{The value of different cues and their integration for action recognition on the UCF101, HMDB51, and J-HMDB datasets (split 1). Adding optical flow and pose is always beneficial. Integration via the proposed Markov chain clearly outperforms the baseline fusion approach. In all cases, the accuracy achieved with estimated optical flow and body parts almost reaches the upper bound performance when providing ground truth values for those inputs.
}
\label{tab:modalities}
\end{table}

\subsection{Action Classification}


Table~\ref{tab:modalities} shows that fusion with the sequential Markov chain model outperforms the baseline fusion consistently across all datasets. The baseline fusion is shown in Figure \ref{baselinefig} and can be considered a strong baseline. It consists of fusing the multiple modalities through feature concatenation followed by a set of fully connected layers. The network is trained jointly. 

Adding pose leads to a substantial improvement over the two-stream version. This confirms that pose plays an important role as complementary modality for action recognition tasks. Again, the Markov chain fusion is advantageous with a large margin. 

For the J-HMDB dataset, ground truth for optical flow and pose is available and can be provided to the method. While not being relevant in practice, running the recognition with this ground truth shows on how much performance is lost due to erroneous optical flow and pose estimates. Surprisingly, the difference between the results is rather small, showing that the network does not suffer much from imperfect estimates. This conclusion can be drawn independently of the fusion method.

Finally, the temporal multi-granularity fusion (MG) further improves results. Especially on HMDB51, there is a large benefit. 



\begin{table}[]
\small
  \begin{center}
    \begin{tabular}{lcccccccccccccccc}
      \toprule

      {} & \multicolumn{3}{c}{Datasets} \\

      \cmidrule{2-4} 
      {Methods} & {UCF101}& {HMDB51}& {J-HMDB}&  \\
      \midrule

     TS Fusion \cite{comp_twofusion}          & 92.5\%           &   65.4\%         & -  \\
     LTC \cite{comp_ltc} 	                  & 91.7\%           &   64.8\%         & -\\
     Two-stream \cite{comp_zisserman}		  & 88.0\%           &   59.4\%         & -\\
     TSN \cite{comp_tsn}	                  & \textbf{94.2}\%  &   69.4\%         & -\\\
     CPD \cite{comp_idtc}		              & 92.3\%           &   66.2\%         & -\\
     Multi-Granular \cite{comp_granular}	  & 90.8\%           &   63.6\%         & -\\
     M-fusion \cite{rw_comb}		          & 89.1\%           &   54.9\%         & -\\
     KVMF \cite{action_kvmf} 		                  & 93.1\%           &   63.3\%         & -\\
     P-CNN \cite{comp_pcnn}		              & -                &-                 & 61.1\%\\
     Action tubes \cite{comp_actiontubes}	  & -                &-                 & 62.5\%\\
     TS R-CNN \cite{comp_tsrcnn} 		      & -                &-                 & 70.5\%\\
     MR-TS R-CNN \cite{comp_tsrcnn} 		  & -                &-                 & 71.1\%\\\hline
     Ours (chained)                    & 91.1\%           & \textbf{69.7\%}  & \textbf{76.1\%}      \\

      \bottomrule
    \end{tabular}\vspace{-4mm}
  \end{center} 
  \caption{Comparison to the state of the art on UCF101, HMDB51, and J-HMDB datasets (over all three splits).}
  \label{tab:comparison}
\end{table}

\subsubsection{Comparison with the state-of-the-art}

Table~\ref{tab:comparison} compares the proposed network to the state of the art in action classificaation. In contrast to Table~\ref{tab:modalities}, the comparison does not show the direct influence of single contributions anymore, since this table compares whole systems that are based on quite different components. Many of these systems also use other features extraction approaches, such as improved dense trajectories (IDT), which generally have a positive influence on the results, but also make the system more complicated and harder to control. Our network outperforms the state of the art on J-HMDB, NTU, and HMDB51. Also, on UCF101 dataset our approach is on par with the current state of the art while it does not rely on any additional hand-crafted features. In two stream case (RGB+OF), if we replace the 3DCNN network by the TSN approach~\cite{comp_tsn}, we obtain a classification accuracy of $94.05\%$ on UCF101 (over 3 splits), which is the state of the art also on this dataset. However, the TSN approach does not allow for action detection anymore. 




\begin{table}[]
\small
  \begin{center}
    \begin{tabular}{lcccccccccccccccc}
      \toprule

      {Methods} & {Cross Subject \%}&  \\
      \midrule

     Deep LSTM \cite{comp_ntu_cvpr} & 60.7\%  \\
      P-LSTM \cite{comp_ntu_cvpr} & 62.93\%   \\
     HOG^2 \cite{comp_ntu_hog} & 32.2\%       \\
     FTP DS \cite{comp_ntu_ftp} & 60.23\%     \\
     ST-LSTM \cite{comp_ntu_lstm} & 69.2\%    \\\hline
         Ours (Pose) & 67.8\%  \\
  Ours (RGB+OF+Pose - Baseline) & 76.9\%  \\
  Ours (RGB+OF+Pose - Chained) & \textbf{80.8}\%  \\

      \bottomrule
    \end{tabular}\vspace{-4mm}
  \end{center} 
  \caption{Comparison to literature on the NTU RGB+D benchmark.}
  \label{tab:comparison}
\end{table}

Finally, we ran the network on the recent NTU RGB+D dataset, which is larger and more challenging than the previous datasets. The dataset is popular for the evaluation of methods that are based on human body pose. Clearly, the result of our network, shown in Table \ref{tab:ntu}, compares favorably to the existing methods. As a result, the used pose estimation network is competitive with pose estimates using depth images and that our way to integrate this information with the raw images and optical flow is advantageous.


\subsubsection{Ordering of modalities in the Markov chain.} 
 
Table \ref{chainorder} shows an analysis on how the order of the modalities affects the final classification accuracy. Clearly, the ordering has an effect. The proposed ordering starting with the pose and then adding the optical flow and the RGB images performed best, but there are alternative orders that do not perform much worse. 

Table~\ref{srefinement} quantifies the improvement in accuracy when adding a modality. 
Clearly, each additional modality improves the results. 

\begin{table}[t]
\footnotesize
\centering
\begin{tabular}{|c|c|c|c|c|c|c|}
\hline
\textbf{Dataset} & \textbf{OPR} & \textbf{ORP}& \textbf{RPO} & \textbf{ROP} & \textbf{PRO}& \textbf{POR}\\ \hline
HMDB51 & 59.8\% & 57.3\% & 54.8\% & 54.1\% & 56.4\% & 60.0\% \\ \hline
UCF101 & 86.8\% & 86.2\% & 84.3\% & 84.7\% & 85.1\% & 87.1\% \\\hline
\end{tabular}\\[-2mm]
\caption{Impact of chain order on the performance (clip accuracy) on UCF101 and HMDB51 datasets (split1). "O" = Optical flow, "P" = Pose and "R" = RGB.}
\label{chainorder}
\end{table}

\begin{table}[t]
\small
\centering
\vspace{-2mm}
\scalebox{0.93}{
\begin{tabular}{|c|c|c|c|}
\hline
\textbf{Dataset} & \textbf{Y\_Pose} & \textbf{Y\_OF}& \textbf{Y\_RGB} \\ \hline
UCF101 & 55.7\% & 83.0\% & 90.4\% \\ \hline
HMDB51 & 40.9\% & 56.4\% & 62.1\% \\\hline
J-HMDB & 47.1\% & 65.3\% & 79.1\% \\\hline
\end{tabular}
}
\caption{Sequential improvement of classification accuracy on UCF101, HMDB51 and J-HMDB datasets (Split1) by adding modalities to the chained network.}
\vspace{-4mm}
\label{srefinement}
\end{table}

\subsubsection{Fusion location}

In principle the chained fusion can be applied to any layer in the network. We studied the effect of this choice. In contrast to the large scale evaluation in Feichtenhofer et al.~\cite{comp_twofusion}, we tested only two locations: FC6 and FC7. Table \ref{fusionLocations} shows a clear difference only on the J-HMDB dataset. There it seems that an earlier fusion, at a level where the features are not too abstract yet, is advantageous. This is similar to the outcome of the study by Feichtenhofer et al.~\cite{comp_twofusion}, where the last convolutional layer worked best. 

\subsubsection{Effect of clip length}


We analyzed the effect of the size of the temporal window on the action recognition performance. Larger windows clearly improve the accuracy on all datasets; see Table \ref{tab:clip_len}. For the J-HMDB dataset (RGB modality) we use a temporal window ranging from $4$ to $16$ frames every $4$ frames. The highest accuracy is obtained with a $16$ frames clip size. Based on the J-HMDB minimum video size, $16$ is the highest possible time frame to be explored. We also tested multiple temporal resolutions for the NTU dataset (pose modality). Again, we obtained the best results for the network with the larger clip length as input. 

The conducted experiments confirm that increasing the length of the clip, we decrease the chance of getting unrelated parts of an action in a video. In addition, with longer sequences, 3D convolutions can better exploit their ability to capture abstract spatio-temporal features for recognizing actions.  

\begin{table}[t]
\small
\centering
\begin{tabular}{|c|c|c|c|}
\hline
\textbf{Fusion Location} & \textbf{UCF101} & \textbf{HMDB51}& \textbf{J-HMDB} \\ \hline
FC7 & 89.8\% & 61.3\% & 73.9\% \\ \hline
FC6 & 89.6\% & 62.1\% & 79.1\% \\\hline
\end{tabular}\vspace{-2mm}
\caption{Classification performance for different fusion locations on UCF101, HMDB51 and J-HMDB datasets (split1).}
\vspace{-2mm}
\label{fusionLocations}
\end{table}

\begin{table}
\small
\centering
\begin{tabular}{|c|c|c|}
\hline
\textbf{Dataset} & \textbf{Clip length} & \textbf{Accuracy} \\ \hline
\multirow{4}{*}{J-HMDB (RGB)} & 4 & 44.8\% \\
  &8 & 49.6\% \\
  &12 & 58.7\% \\
  &16 & 60.8\% \\ \hline
\multirow{2}{*}{NTU RGB+D (Pose)} & 16 & 61.6\% \\
  & 32 & 67.8\% \\\hline
\end{tabular}\vspace{-2mm}
\caption{Effect of the temporal window size. Using more frames as input to the network consistently increases classification performance.}
\label{tab:clip_len}
\end{table}


\subsection{Action Detection}
 
To demonstrate the generality of our approach, we show also results on action detection on UCF101 and J-HMDB. Many of the top performing methods for action classification are not applicable to action detection, because they integrate information over time in a complex manner, are too slow, or are unable to spatially localize the action. 

\begin{figure}[t]
\centering
\includegraphics[width=0.46\textwidth]{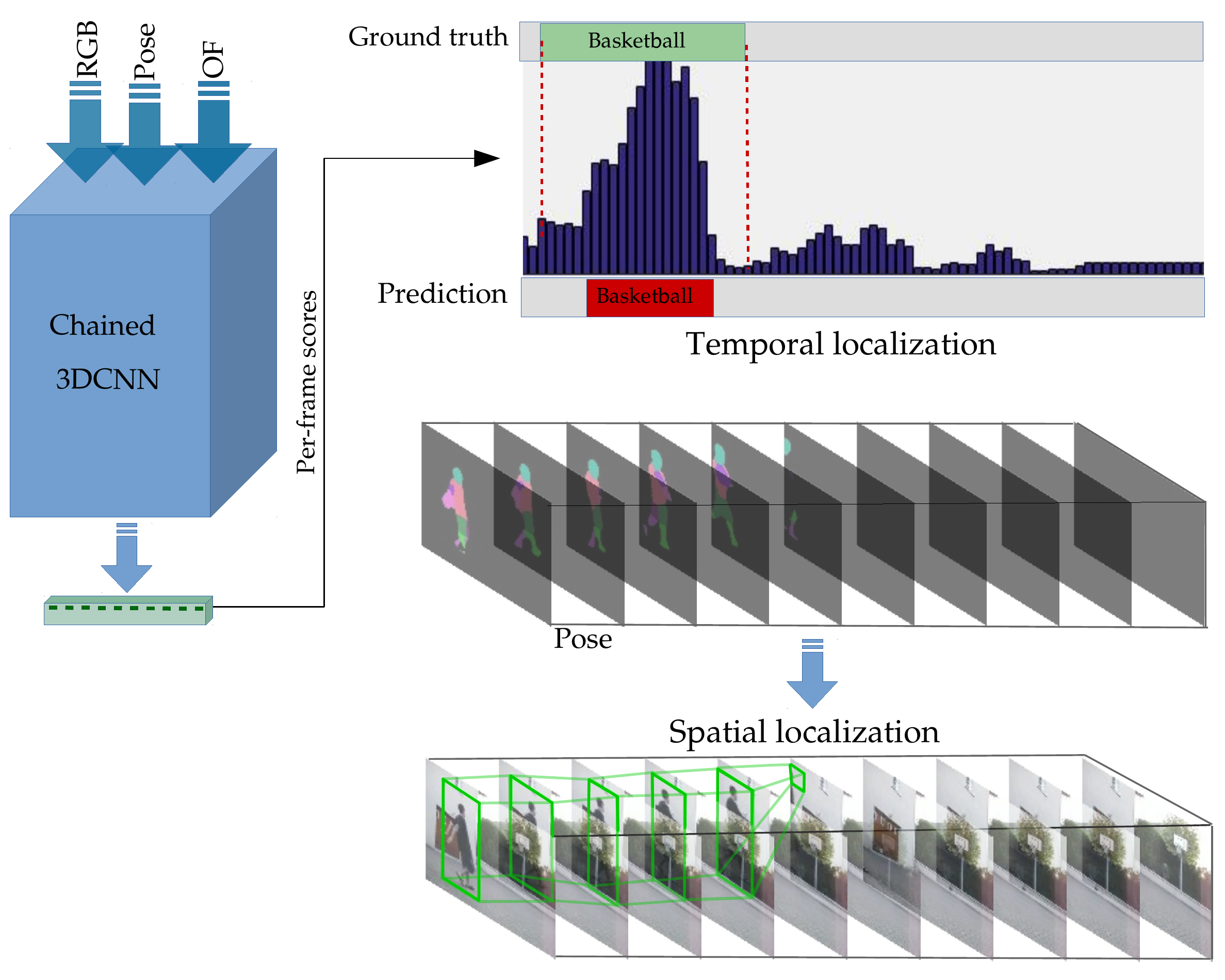}
\caption{Scheme for spatio-temporal action detection. The chained network provides action class scores and body part segmentations per frame. From these we compute action tubes and their actionness scores; see the supplemental material for details.}
\label{cover_detection}
\end{figure}

This is different for our approach, which is efficient and can be run in a sliding window manner over time and provides good spatial localization via the human body part segmentation. In order to create temporally consistent spatial detections, we link action bounding boxes over time to produce action tube \cite{comp_actiontubes}; see the supplemental material for details. We use the frame level action classification scores to make predictions at the tube level. Figure \ref{cover_detection} schematically outlines the detection procedure. 

\begin{figure}[t]
\centering
\includegraphics[width=0.48\textwidth]{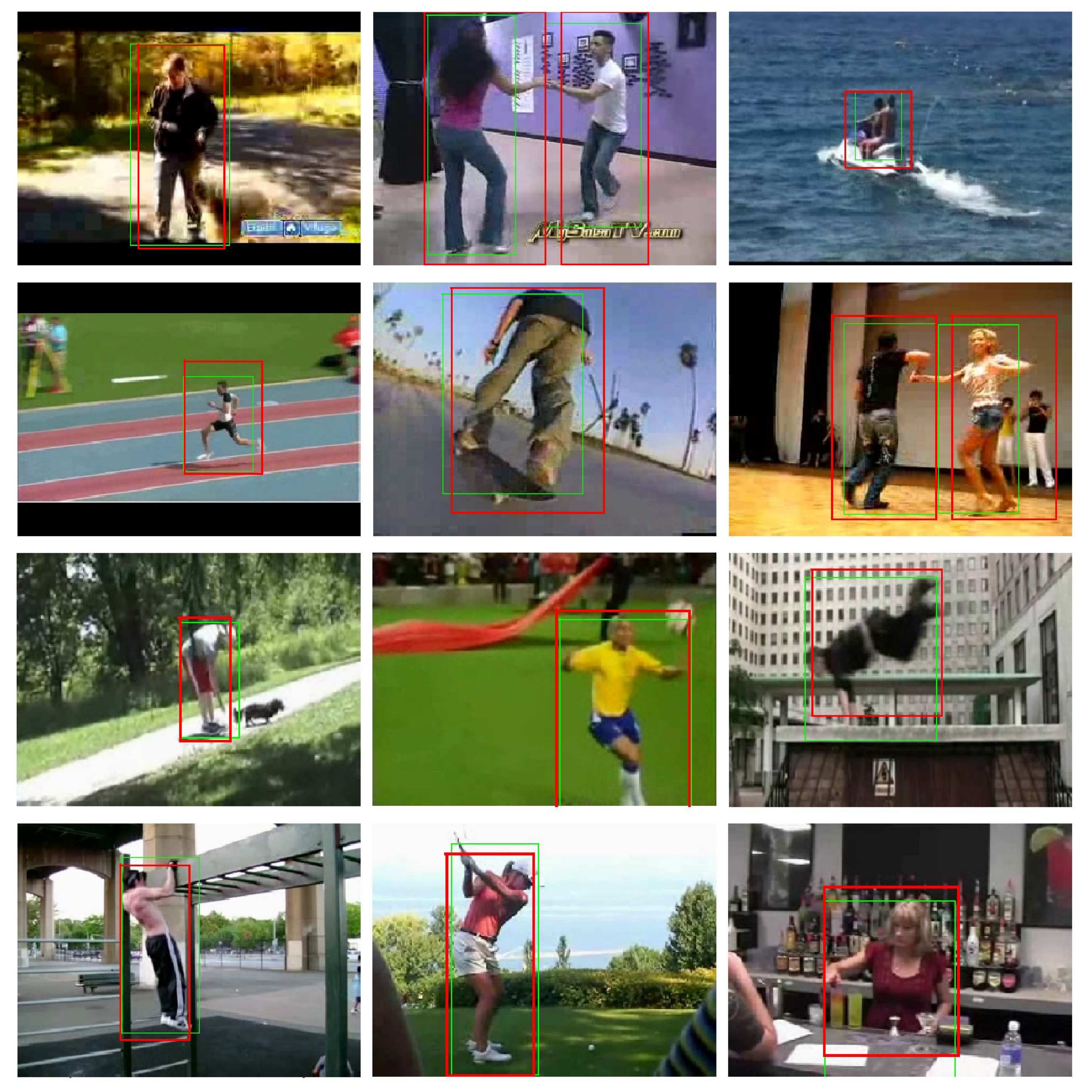}
\caption{ Qualitative results on the action detection task. The first two rows correspond to detections on UCF101, the last ones on J-HMDB. Ground truth bounding boxes are shown in green and detections in red. Our spatial localization is accurate and robust to unusual pose.}
\label{detect_qual}
\end{figure}

We also present a set of qualitative action detection experiments for the UCF and J-HMDB datasets. Figure \ref{detect_qual} shows several examples where we can robustly localize the action, even when unusual pose, illumination, viewpoints and motion blur are presented. Additional results exploring failure cases are provided in supplementary material.     

Following recent works on action detection \cite{comp_actiontubes,iccv15_detect,comp_tsrcnn}, we report video-AP. A detection is considered correct if the intersection over union (IoU) with the ground-truth is above a threshold $\delta$ and the action label is predicted correctly. The IoU between two tubes is defined as the IoU over the temporal domain, multiplied by the average of the IoU between boxes averaged over all overlapping frames. Video-AP measures the area under the precision-recall curve of the action tube predictions.

\renewcommand{\multirowsetup}{\centering} 
\setlength{\tabcolsep}{4pt}
\begin{table}[]
\small
  \begin{center}
    \begin{tabular}{lccccccccccc}
      \toprule

      {} & \multicolumn{5}{c}{J-HMDB} & \\

      \cmidrule{2-6} 
      {IoU threshold ($\delta$)} & {0.1}& {0.2}& {0.3}& {0.4}& {0.5} \\
      \midrule

       Actionness \cite{cvpr16_detect}     & -         &-& -   & -            & 56.4\\
      ActionTubes \cite{comp_actiontubes}	& -          &-& -	 & -	        & 53.3\\
     Weinzaepfel \textit{et al}. \cite{iccv15_detect}		& - &63.1& 63.5& 62.2	 	        & 60.7\\
      Peng \textit{et al}. \cite{comp_tsrcnn}	& -  &74.3& -& -	 	        &73.1 \\ \hline 
      Ours                     & \textbf{78.81}         & \textbf{78.20} & \textbf{77.12} & \textbf{75.05}         & \textbf{73.47}      \\
      \bottomrule
    \end{tabular}\vspace{-3mm}
  \end{center} 
  \caption{Spatial action detection results (Video mAP) on the J-HMDB dataset. Across all IoU thresholds, our model outperforms the state of the art.}
  \label{table:detect_jhmdb}
\end{table}

\renewcommand{\multirowsetup}{\centering} 
\setlength{\tabcolsep}{4pt}
\begin{table}[]
\small
  \begin{center}
    \begin{tabular}{lccccccccccc}
      \toprule

      {} & \multicolumn{4}{c}{UCF101} & \\

      \cmidrule{2-5} 
      {IoU threshold ($\delta$}) & {0.05}&{0.1}& {0.2}& {0.3} \\
      \midrule

      Weinzaepfel \textit{et al}. \cite{iccv15_detect}		& 54.28 & 51.68 &46.77    & 37.82\\
     Yu \textit{et al}. \cite{cvpr15_detect}		& 42.80  &-& -       & -\\  
      Peng \textit{et al}. \cite{comp_tsrcnn}	& 54.46  &50.39& 42.27	 	        &32.70 \\
      Weinzaepfel \textit{et al}. \cite{detect_weakly}	& 62.8  &-& 45.4	 	        &- \\\hline

      \textbf{Ours}                      &\textbf{65.22}         & \textbf{59.52} & \textbf{47.61} & \textbf{38.00}        \\

      \bottomrule
    \end{tabular}\vspace{-4mm}
  \end{center} 
  \caption{Spatio-temporal action detection results (Video mAP) on UCF101 dataset (split1). Across all IoU thresholds, our model outperforms the state of the art.}
  \label{table:detect_ucf}
\end{table}

Table~\ref{table:detect_jhmdb} and Table~\ref{table:detect_ucf} show the video mAP results on spatial and spatio-temporal action detection with different IoU thresholds on J-HMDB and UCF101 (split1) datasets respectively. Although we did not optimize our approach for action detection, we obtain state-of-the-art results on both datasets. Moreover, the approach is fast: spatial detection runs at a rate of 31 fps and spatio-temporal detection with 10 fps. Compared to the recent works \cite{comp_actiontubes,detect_weakly,comp_tsrcnn,cvpr15_detect}, our detection framework has two desirable properties: (1) the pose network directly provides a single detection box per person, which causes a large speed-up; (2) the classification takes advantage of three modalities and the chained fusion, which yields highly accurate per-frame scores.

\section{Conclusion}
\label{sec:conclusion}

We have proposed a network architecture that integrates multiple cues sequentially via a Markov chain model. We have shown that this sequential fusion clearly outperforms other ways of fusion, because it can consider the mutual dependencies of cues during training while avoiding over-fitting due to very large network models. Our approach provides state-of-the-art performance on all four challenging action classification datasets UCF101, HMDB51, J-HMDB and NTU RGB+D while not using any additional hand-crafted features. Moreover, we have demonstrated the value of a reliable pose representation estimated via a fast convolutional network. Finally, we have shown that the approach generalizes also to spatial and spatio-temporal action detection, where we obtained state-of-the-art results as well.

\section{Acknowledgements}

We acknowledge funding by the ERC Starting Grant VideoLearn and the Freiburg
Graduate School of Robotics.


{\small
\bibliographystyle{ieee}

}

\end{document}